\documentclass{article}
\usepackage{spconf,amsmath,graphicx}
\usepackage{bbold}
\usepackage{makecell}
\usepackage{graphics}
\usepackage{multirow}
\usepackage{amssymb}
\usepackage{subcaption}
\usepackage{soul}
\usepackage{graphicx} 
\usepackage[dvipsnames]{xcolor}

\usepackage{url}
\usepackage{multirow}
\usepackage{graphicx}
\usepackage{tabularx}
\newcommand{\uls}[1]{\underline{\smash{}}}

\usepackage{enumitem}
\setlist{nosep, leftmargin=14pt}

\usepackage{mwe} 

\title{Learning to diagnose cirrhosis from radiological and histological labels with joint self and weakly-supervised pretraining strategies}
\name{Emma Sarfati$^{1,2}$ \qquad Alexandre Bône$^{1}$ \qquad Marc-Michel Rohé$^{1}$ \qquad Pietro Gori$^{2}$ \qquad Isabelle Bloch$^{2,3}$}
\address{$^{1}$ Guerbet Research, Villepinte, France \\
     $^{2}$ LTCI, Télécom Paris, Institut Polytechnique de Paris, Saclay, France \\
     $^{3}$ Sorbonne Université, CNRS, LIP6, Paris, France}
\begin{document}
\ninept
\maketitle
\begin{abstract} 

Identifying cirrhosis is key to correctly assess the health of the liver. However, the gold standard diagnosis of the cirrhosis needs a medical intervention to obtain the histological confirmation, \textit{e.g.} the METAVIR score, as the radiological presentation can be equivocal. In this work, we propose to leverage transfer learning from large datasets annotated by radiologists, which we consider as a weak annotation, to predict the histological score available on a small annex dataset. 
To this end, we propose to compare different pretraining methods, namely weakly-supervised and self-supervised ones, to improve the prediction of the cirrhosis. Finally, we introduce a loss function combining both supervised and self-supervised frameworks for pretraining. This method outperforms the baseline classification of the METAVIR score, reaching an AUC of 0.84 and a balanced accuracy of 0.75, compared to 0.77 and 0.72 for a baseline classifier.
\end{abstract}
\section{Introduction}
Cirrhosis diagnosis is important for radiologists, as it can support the differential diagnosis of liver masses, such as hepatocellular carcinoma, a liver primary cancer \cite{cirrhosishcc}. The presence - or absence - of cirrhosis can more generally be a signal for hepatic pathologies. In clinical routine, the diagnosis of cirrhosis can be performed with three different approaches. First, the gold standard method is the histological analysis obtained by biopsy (or following a resection). However, this method is clinically invasive, hence risky and expensive. Secondly, a clinical examination is done to detect the  potential signs of a terminal cirrhosis stage (an interrogation about alcohol consumption, visible signs of jaundice or ascites, related to a swollen belly). Thirdly, CT-scans in portal venous phase can be analyzed by radiologists to find imaging features of the disease, but the task remains difficult and mainly uneven as the diagnosis can change from one radiologist to another, with typically low inter-rate agreement scores \cite{fibrosis}. 

Several methods have been proposed for automatic cirrhosis prediction from medical images. These methods use mainly Deep Convolutional Neural Networks (DCNN) for cirrhosis prediction, considered as a classification or a regression problem \cite{yasaka}. While some methods use large backbones with millions of parameters as encoder such as DenseNet-121 or ResNet-Inception-v2~\cite{li,byra}, lighter networks have proved to provide very good results in terms of accuracy \cite{yin}. However, these state-of-the-art methods rely on histopathological diagnosis as label features \cite{li}, using the METAVIR score classification corresponding to different stages of fibrosis (F0/F1/F2/F3/F4), or the Inuyama score classification, which is close to the METAVIR one. The majority of these studies use large labeled datasets with hundreds or thousands of histologically-diagnosed patients \cite{choi}. Replicating these approaches requires large volumes of images with corresponding biopsy, which are difficult to obtain. By contrast, obtaining large volumes of CT-scans without annotations is much easier, and getting \textit{a posteriori} annotations from radiologists is still possible because no medical intervention is needed. 
To cope with the limited data availability, deep learning studies have demonstrated the possibility of advantageously pre-training models on large databases to prepare their transfer on a deployment database~\cite{lwf}. Pretraining databases can be labeled (and used in transfer learning \cite{supcon}), unlabeled (used in self-supervised learning, for instance with SimCLR \cite{simclr}), or weakly-labeled like in \cite{yaware}, \textit{i.e.} annotated with a label that is close to the reference one, and hence that can be regarded as a proxy for the latter. Pre-training can be beneficial when there are few labeled images and many unlabeled (or weakly/noisily labeled) images.

In this work, we propose to explore several pretraining methods to improve the prediction of a binarized METAVIR score from a small CT-scan dataset, using a large weakly-labeled CT-scan dataset. 
We compare three different approaches; first, we explore the effect of a standard transfer learning method to improve the prediction of the METAVIR score, \textit{i.e.} we pretrain a supervised model on a weak (or noisy) label and then re-use the weights to predict the strong label. The second part presents the impact of self-supervised pretraining (SimCLR \cite{simclr}) for the same purpose. Finally, we study the introduction of the radiological label within the self-supervised framework, first using the existing Supervised Contrastive Learning model (SupCon, \cite{supcon}), then enhancing the latter by proposing a weighted sum of SimCLR and CrossEntropy loss functions. In this article, we denote the radiological labels as ``weak" labels, as we suppose that they can be seen as noisy approximations of the reference histological labels.

\section{Method}
\paragraph*{Dataset.} Two datasets are leveraged in this study. First, $\mathcal{D}_{histo}$, contains 106 CT-scans from different patients in portal venous phase, with an identified histopathological status obtained by a histological analysis, designated as $Y_{histo}$. The latter is binarized to indicate the absence or presence of advanced fibrosis \cite{li} obtained by the separation F0/F1/F2 vs. F3/F4. The pathological class contains 78 patients while the healthy one includes 28 patients. The second dataset, $\mathcal{D}_{radio}$, consists of 2,799 CT-scans of patients in portal venous phase with a radiological annotation, \textit{i.e} realized by a radiologist, indicating four different stages of cirrhosis: no cirrhosis, mild cirrhosis, moderate cirrhosis and severe. We also binarize this label to obtain no cirrhosis versus mild/moderate/severe ($Y_{radio}$). This dataset contains 919 pathological (supposedly cirrhotic, i.e. mild/moderate/severe) subjects and 1,880 healthy ones. 

All images have a 512x512 size, and we clip the intensity values between -100 and 400. We work with 2D slices rather than 3D volumes. Moreover, we select the slices based on the liver segmentation of the patients. We keep the top 70\% most central slices with respect to automatically-computed liver segmentation maps.

\begin{figure}[h!]
    \centering
    \includegraphics[width=8.5cm]{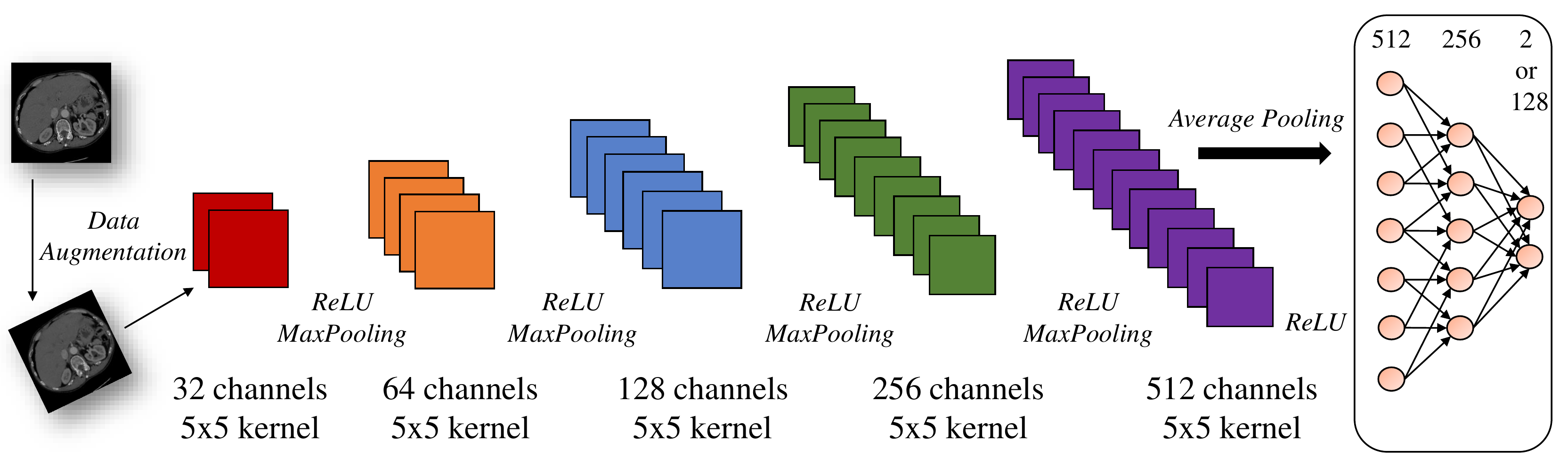}
    \caption{The DCNN used in our method.}
    \label{net}
\end{figure}

\paragraph*{Architecture and optimization.} ~\\\\
\textit{Backbone.} \quad Inspired by \cite{yin}, we propose a baseline backbone based on a simple architecture, illustrated in Figure~\ref{net}. The 512x512 input images are passed through five convolutional layers, each of them using a 5x5 kernel, followed by a ReLU activation function before a  2x2 MaxPooling operation. The first layer has 32 channels, and the number of channels is doubled at every layer. It then generates a 512x20x20 feature map that is pool-averaged, ending in a 512-dimensional flat vector. Two dense layers end the network, mapping the 512-dimensional vector to a 256-dimensional vector, then to a binary output followed by a softmax. For self-supervised learning, we replace the last linear layer by another one with an output dimension of 128 to be consistent with the original SimCLR method \cite{simclr}.  We denote by $f(.)$ the backbone encoder preceding the dense layers, and $p(.)$ the projector composed of the two dense layers. We obtain either a 128-dimensional or a 2-dimensional output, $z=p(f(x))$. This architecture is used as a basis for all our experiments.  
\vspace*{-\baselineskip}
\noindent \\
\noindent\textit{Sampling strategy and loss functions.} \quad
For the baseline performance, the proposed encoder (Figure~\ref{net}) is learned on $\mathcal{D}_{histo}$. In a preliminary experiment, the ratio of leveraged training data is artificially reduced from 100\% to 80\%, 60\% and 40\% in order to assess the impact of the number of cases on classification performance. Then, four pretraining experiments are led to improve the baseline performance for cirrhosis classification. 

First, we train a supervised model (backbone architecture, Figure~\ref{net}) to predict the binarized radiological labels present in $\mathcal{D}_{radio}$, which we consider as a weak label for the gold standard $Y_{histo}$, and then use a transfer learning strategy to predict the binarized histological labels in $\mathcal{D}_{histo}$. This first pretraining experiment can be regarded as weakly-supervised for our purpose, hence we denote the binary cross entropy used there as $\mathcal{L}_{weak}$.

As a second experiment, we leverage a self-supervised pretraining approach, SimCLR, using the original NTXentLoss \cite{simclr,simclr2}: 
\begin{equation*}
    \mathcal{L}_{SimCLR}=-\frac{1}{2N}\sum_{i=1}^{2N}\log\frac{\exp(sim(z_i,z_{j(i)})/\tau)}{ \sum_{k=1}^{2N} 1[k\neq i] \exp(sim(z_i,z_k)/\tau)}
\end{equation*}
with $j(i)$ denoting the positive with respect to $i$, \textit{i.e.} the second augmented version of the original image $x_i$, $N$ denoting the batch size and $z_i=p(f(x_i))$, $z_{j(i)}=p(f(x_{j(i)}))$ denote the output vectors of the image augmentations $x_i$ and $x_{j(i)}$ passed through two random augmentations modules, $\mathcal{T}$ and $\mathcal{T}'$. The similarity is defined as $sim(z_i,z_j)=z_i^\top z_j$. We fix the temperature parameter at $\tau=0.1$. 

For the third experiment, we explore SupCon \cite{supcon} using $Y_{radio}$ to pair samples from the same class together:
\begin{equation*}
{\footnotesize
\resizebox{\columnwidth}{!}{$\displaystyle
     \mathcal{L}_{SupCon}=\sum_{i=1}^{2N}\frac{-1}{|P(i)|}\sum_{j\in P(i)}\log\frac{\exp(sim(z_i,z_j)/\tau)}{ \sum_{k=1}^{2N} \mathbb{1}[k\neq i] \exp(sim(z_i,z_k)/\tau)} $
}
}
\end{equation*} 
where $P(i)$ denotes the set of all the indices of samples belonging to the same class as the input image. 

\begin{figure*}[h!]
    \centering
    \includegraphics[width=12.4cm,height=6.9cm]{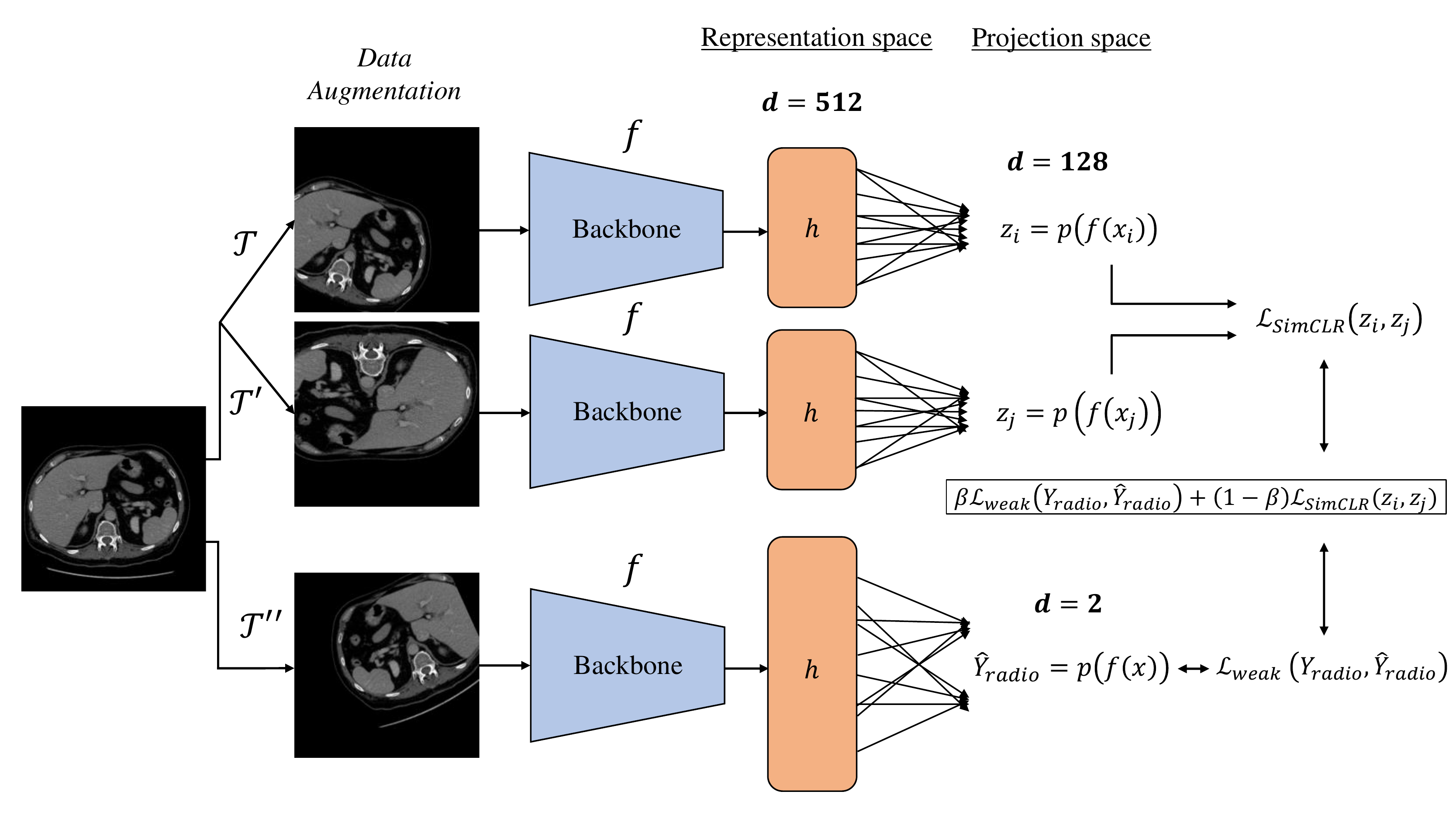}
    \caption{The proposed pretraining strategy, \textit{weak-SimCLR}, which loss function is a sum of the binary cross-entropy computed with $Y_{radio}$ ($\mathcal{L}_{weak}$) and of the SimCLR loss function ($\mathcal{L}_{SimCLR}$).}
    \label{global}
\end{figure*}

Fourth, to maximize the potential information given by $Y_{radio}$ as well as the representation power offered by SimCLR, we propose a new loss function for pre-training, $\mathcal{L}_{weak-SimCLR}$, 
a simple weighted sum of the binary cross entropy and the NTXentLoss:
\begin{equation}
    \mathcal{L}_{weak-SimCLR}=\beta \mathcal{L}_{weak}+(1-\beta)\mathcal{L}_{SimCLR}
\end{equation}
where $\beta \in [0,1]$ is an hyper-parameter. To compute this function, the only change in the training process is that the original image is passed through a third data augmentation module $\mathcal{T}''$, before being passed to the backbone and two dense layers, mapping the 512-dimensional representation vector to a 256-dimensional then to a 2-dimensional one. Note that all the weights are shared between the supervised and unsupervised branches, only the last dense layers, due to the difference in output dimensions, differ between both (see Figure~\ref{global}). 

Finally, for sampling, we observe a class imbalance which we try to fix using a weighted sampling during training, and we report the results of the balanced accuracy scores. As we work with 2D slices rather than 3D volumes, we compute the average probability of having the pathology per patient. The evaluation results presented later are based on the patient-level aggregated prediction.
\\

\noindent\textit{Data augmentation and optimization setting.} \quad Unsupervised contrastive learning methods such as SimCLR \cite{simclr} typically require heavy data augmentations on input images, in order to strengthen the association between positive samples in the representation space \cite{augmentations}. In our work, we leverage three specific types of augmentations: crops, flips and rotations. During our experiments, we also inspected the effect of CutOut \cite{cutout}, which proved not to increase the performances of our models. Data augmentations are computed on the GPU, using the Kornia library \cite{kornia}. During inference, we remove the augmentation module to only keep the original input images.

We run our experiments on a Tesla V100 with 16GB of RAM and a 6 CPU cores and we used the PyTorch-Lightning library to implement our models. All the models share the same random data augmentation module, with a batch size of $N=92$ and a fixed number of epochs $n_{epochs}=200$. For the pretraining experiments, \textit{i.e.} the models trained on the large dataset $\mathcal{D}_{radio}$, we fix a learning rate (LR) of $\alpha=10^{-4}$  and a weight decay of $\lambda=10^{-4}$. For the classification experiments, \textit{i.e.} the models trained on the small dataset $\mathcal{D}_{histo}$, we fix a learning rate (LR) of $\alpha=10^{-5}$  and a weight decay of $\lambda=10^{-3}$. For all the experiments, we added a cosine decay learning rate scheduler. Finally, we fix the hyper-parameter $\beta$ of \textit{weak-SimCLR} at 0.5, unless otherwise specified.

\paragraph*{Evaluation protocol.}  We evaluate our methods using two different procedures. First, we extract the 512-dimensional vectors $f(x)$ from the representation space (see Figure~\ref{global}) and train a simple logistic regression on the frozen representations with a default regularization parameter of $\lambda=1$, using scikit-learn. This procedure can be thought as a linear evaluation in the representation space. Hence, we can denote this first evaluation method a ``cross-validated (CV) linear evaluation", as introduced in Table~\ref{results}. Secondly, we fine-tune the whole network initializing the latter with the pretrained weights. For both evaluation procedures, we validate the results with a stratified 5-fold cross-validation.
\section{Results}
\label{sec:results}
\begin{figure}[h!]
    \centering
    \includegraphics[width=5.5cm]{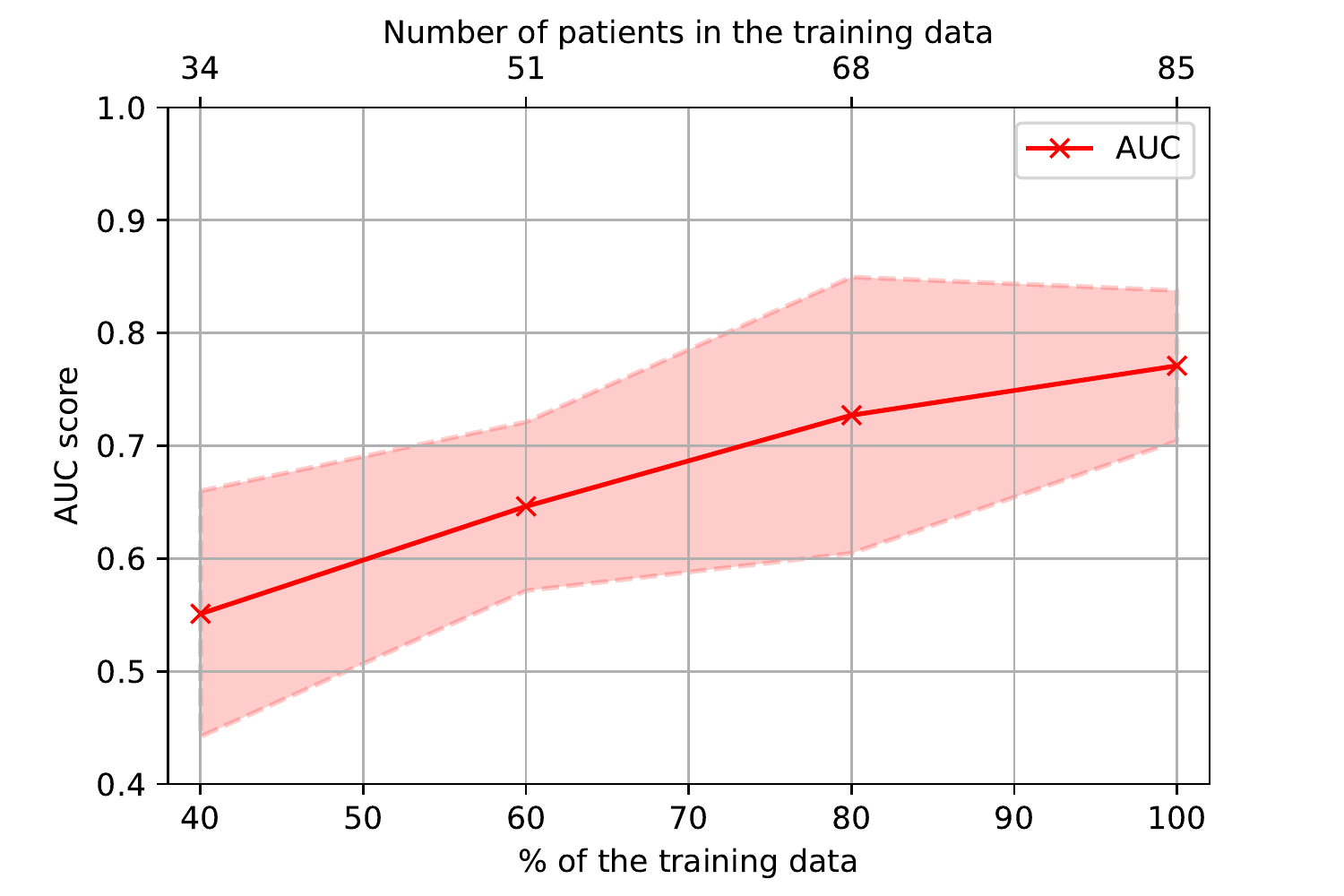}
    \caption{Evolution of the averaged cross-validation AUC with respect to the percentage of the training data.}
    \label{trainingdata}
\end{figure}
We first report in Figure~\ref{trainingdata} the evolution of the averaged cross-validation AUC with respect to the percentage of the training data kept during training, as well as the associated standard deviations. As a reference, \cite{yasaka, yin} reported respectively an AUC of 0.76 for 186 patients and 0.89 for 202 patients in their training sets, while $\mathcal{D}_{histo}$ presents 85 patients for training and 21 for validation.
\vspace*{-\baselineskip}
\paragraph*{Transfer learning results.}
\begin{table}[h!]
\resizebox{\columnwidth}{!}{%
\centering
\begin{tabular}{|c||cc|cc|}
\hline
\multirow{2}{*}{\begin{tabular}[c]{@{}c@{}}\\Pretraining\\ method\end{tabular}} & \multicolumn{2}{c|}{AUC}                                 & \multicolumn{2}{c|}{Balanced Accuracy}                                  \\ \cline{2-5} 
 &
  \multicolumn{1}{c|}{\begin{tabular}[c]{@{}c@{}}CV linear \\ evaluation\end{tabular}} &
  Fine-tuning &
  \multicolumn{1}{c|}{\begin{tabular}[c]{@{}c@{}}CV linear \\ evaluation\end{tabular}} &
  Fine-tuning \\ \hline
None                   & \multicolumn{1}{c|}{/}                & 0.77 ($\pm$0.07) & \multicolumn{1}{c|}{/}                & {0.72 ($\pm$0.07)} \\ \hline
Supervised             & \multicolumn{1}{c|}{0.64 ($\pm$0.19)} & 0.65 ($\pm$0.10) & \multicolumn{1}{c|}{0.56 ($\pm$0.10)} & 0.64 ($\pm$0.09)                \\ \hline
SimCLR                 & \multicolumn{1}{c|}{0.78 ($\pm$0.09)} & 0.78 ($\pm$0.11) & \multicolumn{1}{c|}{0.68 ($\pm$0.04)} & 0.69 ($\pm$0.08)                \\ \hline
SupCon                 & \multicolumn{1}{c|}{0.61 ($\pm$0.05)} & 0.65 ($\pm$0.13) & \multicolumn{1}{c|}{0.59 ($\pm$0.05)} & 0.67 ($\pm$0.14)                \\ \hline
Ours &
  \multicolumn{1}{c|}{{\underline{\smash{ \textbf{0.84 ($\pm$0.05)}}}}} &
  {\underline{\smash{ \textbf{0.81 ($\pm$0.03)}}}} &
  \multicolumn{1}{c|}{\underline{\smash{\textbf{0.75 ($\pm$0.06)}}}} &
  \underline{\smash{\textbf{0.73 ($\pm$0.08)}}} \\ \hline
\end{tabular}}%
\caption{Results AUCs and balanced accuracies of our experiments, with a value of $\beta=0.5$ for \textit{weak-SimCLR}. Best results by column are \textbf{\underline{\smash{underlined}}}. The standard deviations come from the AUCs within the folds and the AUCs are averaged by fold.}
\label{results}
\end{table}
Table~\ref{results} gathers the AUC and balanced accuracy performance measures for each transfer learning configuration. It first provides results of the supervised classifier directly trained on $\mathcal{D}_{histo}$, from random initial weights, which is the baseline result. Next, it presents the performances of the four pretraining experiments that were led, ending with the proposed model, \textit{weak-SimCLR}. First, it shows that both SimCLR and \textit{weak-SimCLR} overcome the baseline AUC. In particular, \textit{weak-SimCLR} presents an increase of 7\% in cross-validated linear evaluation, and 5\% in fine-tuning, which confirms that the representation space built by the proposed model can provide a globally relevant separation. It can be noted that the first evaluation method, training a logistic regression on the frozen representation vectors previously trained on $\mathcal{D}_{radio}$, is faster and less computationally demanding than the full fine-tuning. For the accuracy, the proposed method slightly outperforms the baseline score, in fine-tuning and transfer learning with respectively 0.75 and 0.73 for balanced accuracy scores, compared to 0.72 for the supervised classifier.
\paragraph*{Robustness study.}
\begin{table}[h!]
\resizebox{\columnwidth}{!}{%
\begin{tabular}{|c||cc|cc|}
\hline
\multirow{3}{*}{$\beta$} &
  \multicolumn{2}{c|}{AUC} &
  \multicolumn{2}{c|}{Balanced Accuracy} \\ \cline{2-5} 
 &
  \multicolumn{1}{c|}{\begin{tabular}[c]{@{}c@{}}CV linear \\ evaluation\end{tabular}} &
  Fine-tuning &
  \multicolumn{1}{c|}{\begin{tabular}[c]{@{}c@{}}CV linear \\ evaluation\end{tabular}} &
  Fine-tuning \\ \cline{2-5} 
 &
  \multicolumn{1}{c|}{/} &
  0.77 ($\pm$0.07) &
  \multicolumn{1}{c|}{/} &
  {0.72 ($\pm$0.07)} \\ \hline
0 (SimCLR) &
  \multicolumn{1}{c|}{0.78 ($\pm$0.09)} &
  0.78 ($\pm$0.11) &
  \multicolumn{1}{c|}{0.68 ($\pm$0.04)} &
  0.69 ($\pm$0.08) \\ \hline
0.2 &
  \multicolumn{1}{c|}{0.75 ($\pm$0.08)} &
  0.78 ($\pm$0.10) &
  \multicolumn{1}{c|}{0.67 ($\pm$0.05)} &
  0.73 ($\pm$0.13) \\ \hline
0.4 &
  \multicolumn{1}{c|}{0.82 ($\pm$0.10)} &
  0.81 ($\pm$0.07) &
  \multicolumn{1}{c|}{0.71 ($\pm$0.09)} &
  \underline{\smash{ \textbf{0.75 ($\pm$0.07)}}} \\ \hline
0.5 &
  \multicolumn{1}{c|}{{\underline{\smash{ \textbf{0.84 ($\pm$0.05)}}}}} &
  {\underline{\smash{ \textbf{0.81 ($\pm$0.03)}}}} &
  \multicolumn{1}{c|}{{\underline{\smash{ \textbf{0.75 ($\pm$0.06)}}}}} &
  0.73 ($\pm$0.08) \\ \hline
0.8 &
  \multicolumn{1}{c|}{0.74 ($\pm$0.06)} &
  0.79 ($\pm$0.09) &
  \multicolumn{1}{c|}{0.68 ($\pm$0.07)} &
  0.66 ($\pm$0.11) \\ \hline
1 (Supervised) &
  \multicolumn{1}{c|}{0.64 ($\pm$0.19)} &
  0.65 ($\pm$0.10) &
  \multicolumn{1}{c|}{0.56 ($\pm$0.10)} &
  0.64 ($\pm$0.09) \\ \hline
\end{tabular}%
}
\caption{Results AUCs and balanced accuracies of the proposed model, \textit{weak-SimCLR}, with respect to the value of $\beta$ (supervision) incroporated in the model. Best results by column are \textbf{\underline{\smash{underlined}}}.}
\label{beta}
\end{table}
Table~\ref{beta} gathers the AUC and balanced accuracy performance measures for each value of $\beta$ of the proposed method, \textit{weak-SimCLR}. For the cross-validated linear evaluation, the value of $\beta=0.5$ reaches the best AUC and balanced accuracy performances. For the AUC, it can be noted that SimCLR and both $\beta=0.4$ and $\beta=0.5$ of \textit{weak-SimCLR} outperform the baseline score of 0.77. Moreover, all the tested values of $\beta$, except 0 and 1, provide a better AUC result for fine-tuning. In terms of accuracy, only the value of $\beta=0.5$ gives a better result than the baseline score with a gain of $3\%$. For the fine-tuning part, the scores are stable with respect to the CV linear evaluation ones. 

\section{Discussion and Conclusion}
We explored four pretraining methods to improve the cirrhosis classification on a small histologically-labeled dataset, using a large radiologically-labeled dataset.
We proposed a method to improve the automatic diagnosis of cirrhosis, using radiological (weak) annotations for pre-training models. To the best of our knowledge, no other work using radiological annotation to improve the automatic prediction of the histological one for cirrhosis prediction with deep learning has been proposed in the literature. The proposed method relies on the combination of weakly-supervised and self-supervised approaches, with the share of each model being a hyper-parameter that can be tuned. It improved the automatic diagnosis of cirrhosis, based on both AUCs and balanced accuracies. Notably, it outperforms two state-of-the-art methods in contrastive learning \cite{simclr,supcon}, as well as the baseline classification realized on our small histologically labeled dataset. We obtain a competitive performance with respect to the literature \cite{yin}, reaching an AUC of 0.84, with only 106 annotated CT-scans. 

The first limit of our work is the baseline performance which may be improved by testing different backbones, other than the one we proposed (see Figure~\ref{net}), such as ResNet-50. The self-supervised block in the proposed model could also be tested with other contrastive or non-contrastive methods (\textit{e.g.} SimSIAM \cite{simsiam} or BYOL \cite{byol}). Second, it could be interesting to evaluate our method on an external public dataset such as the Liver Hepatocellular Carcinoma (LIHC) dataset from the Cancer Genome Atlas \cite{lihc}. Finally, more evaluation procedures could be added to assess the robustness of the proposed strategy regarding the small number of patients in $\mathcal{D}_{histo}$. Indeed, the standard deviations in our results are important, due to some disparities between the folds. K-fold cross validation with higher values of K could be tested, or even Leave-One-Out cross-validation.


\paragraph*{Compliance with ethical standards.} This research study was conducted retrospectively using human data collected from various medical centers, whose Ethics Committees granted their approval. Data was de-identified and processed according to all applicable privacy laws and the Declaration of Helsinki.
\paragraph*{Acknowledgments.} This work was supported by Région Île-de-France (ChoTherIA project) and ANRT (CIFRE \#2021/1735).


\bibliographystyle{IEEEbib}
\bibliography{strings,refs}

\end{document}